\pdfoutput=1
\documentclass[11pt]{article}

\IfFileExists{acl.sty}{
  \usepackage[preprint]{acl}
}{
  \let\scchartalignbibliographystyle\bibliographystyle
  \renewcommand{\bibliographystyle}[1]{\scchartalignbibliographystyle{latex/##1}}
  \usepackage[review]{latex/acl}
  \let\bibliographystyle\scchartalignbibliographystyle
}
\usepackage{times}
\usepackage{latexsym}
\usepackage[T1]{fontenc}
\usepackage[utf8]{inputenc}
\usepackage{microtype}
\IfFileExists{inconsolata.sty}{\usepackage{inconsolata}}{}
\usepackage{amsmath,amssymb,amsfonts}
\usepackage{subcaption}
\usepackage{graphicx}
\graphicspath{{../}{../figures/}{figures/}}
\usepackage{xcolor}
\usepackage{multirow}
\usepackage{booktabs}
\usepackage{pifont}
\usepackage{array}
\usepackage{makecell}
\usepackage{enumitem}

\newcommand{\cmark}{\textcolor{green!60!black}{\ding{51}}}
\newcommand{\xmark}{\textcolor{red!70!black}{\ding{55}}}
\newcommand{\pmark}{\textcolor{orange!85!black}{$\triangle$}}

\begin{document}

\title{SciChartAlign: Scientific Chart-Message Aligned Dataset and Applications}

\author{
  Hyeonseok Kang \quad Hyuk Namgoong \quad Jaewon Heo \quad Sangkeun Jung\thanks{Corresponding author.} \\
  Department of Computer Science and Engineering, Chungnam National University \\
  \texttt{dnfldjaak11@gmail.com}, \texttt{hyuk199@gmail.com}, \\
  \texttt{jwheo2hyw@gmail.com}, \texttt{hugmanskj@gmail.com}
}

\maketitle

\begin{abstract}
A scientific chart both represents experimental data and serves as evidence for the message an author draws from it, so a chart, the evidence needed to reconstruct it, and the author-written finding it supports form one unit. Existing resources align only part of that unit: a published chart is paired with a caption, a data table, or code, while the finding it supports is left outside the instance. To align the unit as a whole, we introduce SciChartAlign, a dataset of 4,342 charts from 2,336 published NLP papers that aligns each source chart and caption with a reconstructed data table, executable Python code, rendered chart, visual style metadata, and an author-written finding. Paper-native components are kept distinct from metadata proposed by three multimodal large language models and curated by human experts for visual fidelity and chart--finding consistency. External evaluation reproduces the curated rendered chart with 0.812 accuracy, and 83.19\% of non-selected candidate findings are semantically covered by the curated finding. Across chart data extraction, chart explanation, and retrieval-augmented chart-to-code generation, dataset examples improve chart-to-table extraction and caption-style explanation, and retrieving findings together with code attains the highest LLMScore (0.7624) and the lowest execution error rate (0.1295). SciChartAlign thus aligns reproducible visual evidence with author-written messages in published NLP papers.
\end{abstract}

\begin{figure}[t]
  \centering
  \begin{subfigure}[t]{0.5\columnwidth}
    \centering
    \includegraphics[width=\linewidth,height=7.3cm,keepaspectratio]{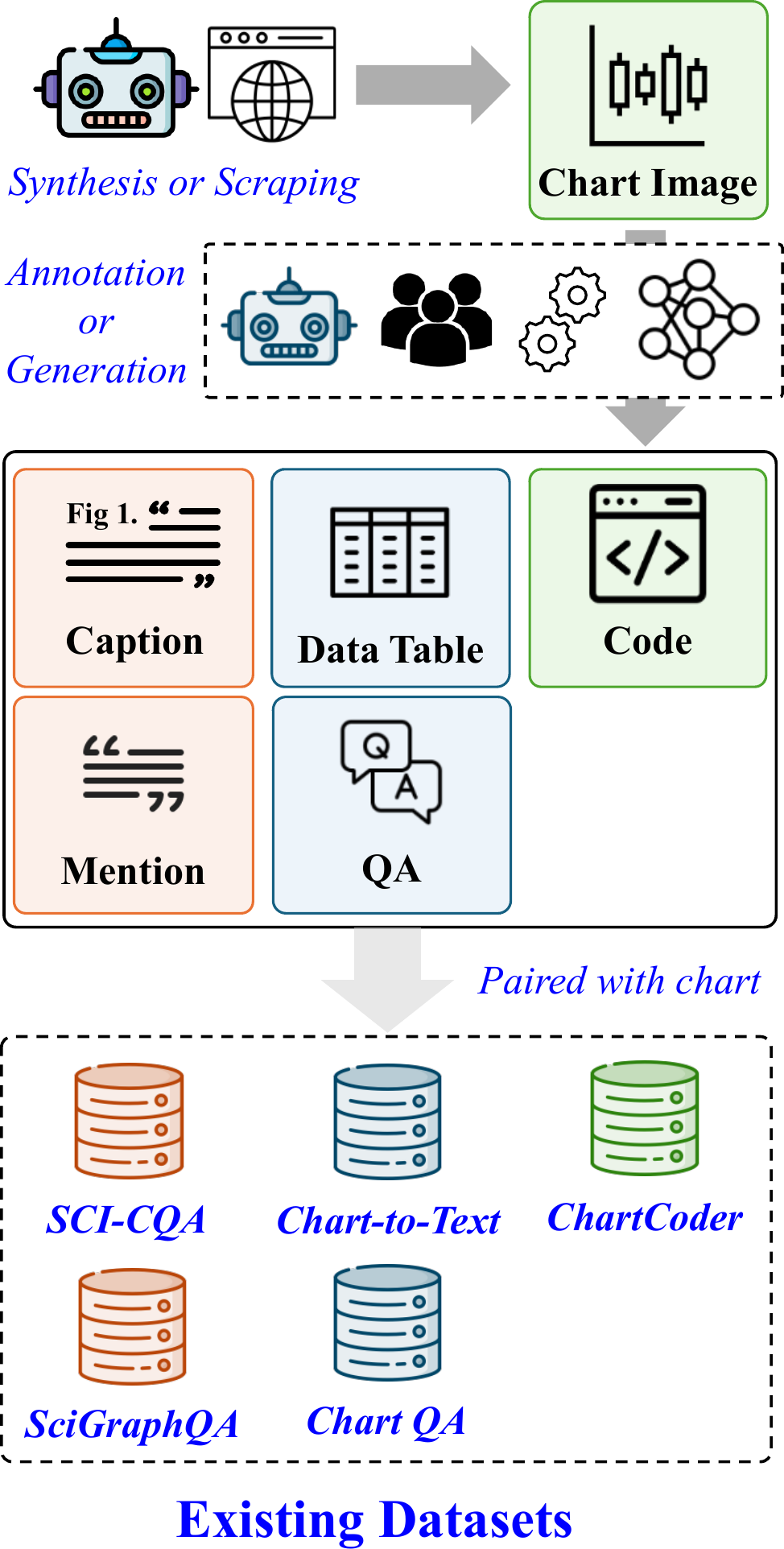}
    \caption{Chart-centric dataset}
    \label{fig:overview_a_arr}
  \end{subfigure}\hfill
  \begin{subfigure}[t]{0.5\columnwidth}
    \centering
    \includegraphics[width=\linewidth,height=7.3cm,keepaspectratio]{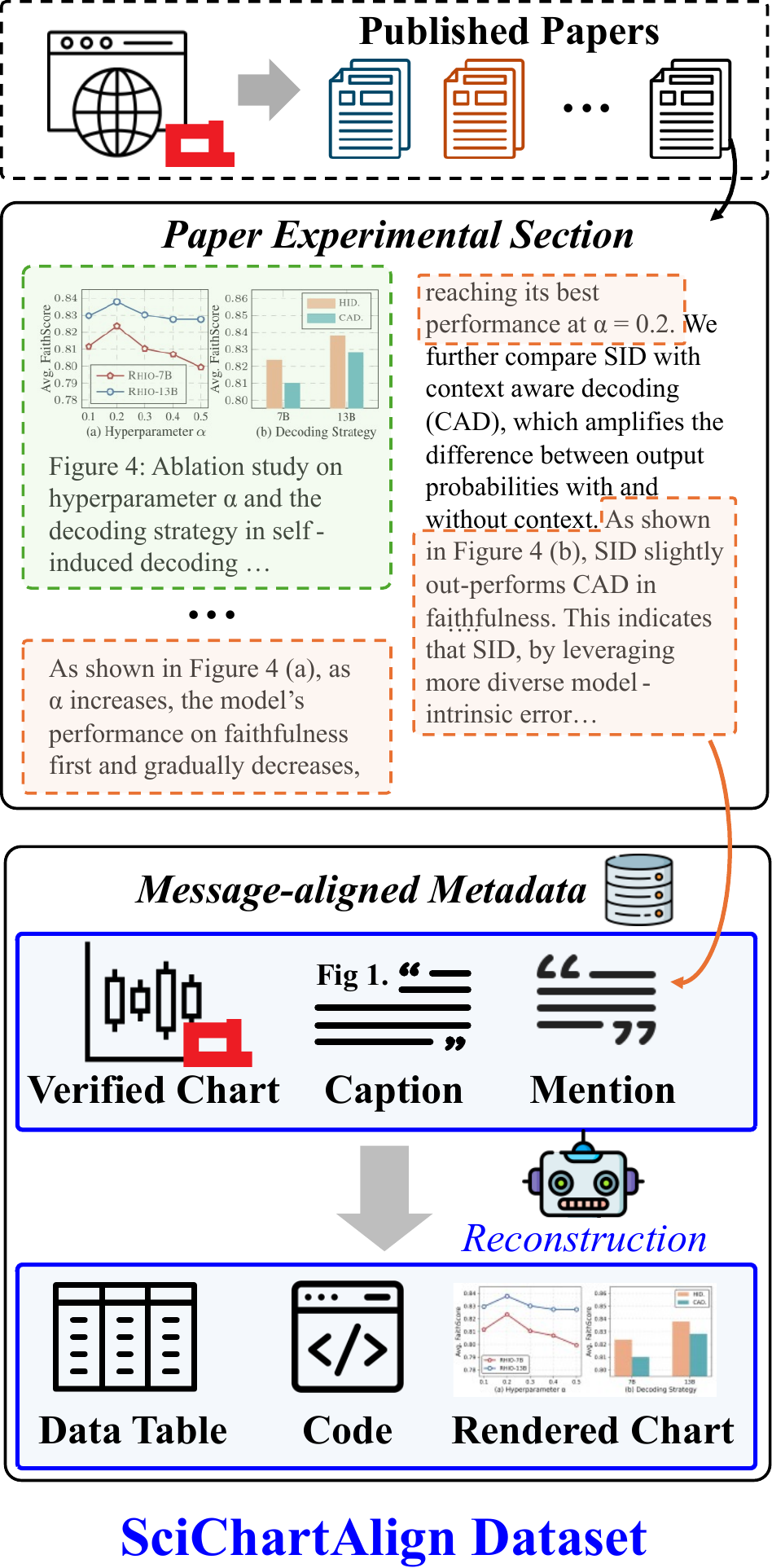}
    \caption{Message-aligned dataset}
    \label{fig:overview_b_arr}
  \end{subfigure}
  \caption{Components of chart datasets. (a) A chart-centric dataset pairs a chart with QA, captions, data tables, or code. (b) The message-aligned dataset (ours) links a published chart to reconstructed data, executable code, rendered output, visual style, caption, and author-written finding.}
  \label{fig:overview_arr}
\end{figure}

\section{Introduction}

Charts in scientific papers carry experimental evidence. Authors use them to organize model comparisons, distributions, performance trends, and ablation results, and figures serve as a major channel for presenting complex results concisely~\cite{divecha2023tables}. Their value depends on whether the visual encoding preserves the relevant relationships, since poor visualizations introduce misinterpretation and reduce trust~\cite{franconeri2021science}.

A chart in a paper is not read in isolation. Its caption identifies the displayed variables and conditions, while the surrounding experimental text states the comparison, trend, performance difference, or interpretation that the chart is presented to support. The chart and that author-written message form one evidential unit: the message tells a reader which relation in the plot carries the result, and the plot is what licenses the message. Reproducing a scientific chart therefore means reproducing not only its marks and values but also the relation it was drawn to make visible.

Existing resources do not preserve this unit. Chart question answering and summarization pair charts with questions or descriptions, data extraction pairs images with tables, and chart-to-code generation pairs images with plotting programs~\cite{chartqa,Chart-to-Text,chartcoder}. A chart produced from such supervision can be structurally correct while emphasizing nothing in particular, because no field records what the chart was meant to show. Resources that do retain author-written text are limited in the rest of the tuple: as summarized in Figure~\ref{fig:overview_arr} and Table~\ref{tab:dataset_comparison_arr}, those linking a chart to a finding provide neither executable code nor a reconstruction-oriented data table, and no existing resource aligns code, chart image, data, and message in one validated instance. We ask: \textit{Can published scientific charts be represented as aligned units that preserve both reproducible visual evidence and the author-written message they support?}

To answer it, we construct \textbf{SciChartAlign} (Scientific-paper Chart--Message Alignment), a chart dataset derived from 2025 main-track long papers at ACL, EMNLP, NAACL, and COLING. Each validated instance aligns the source chart and caption with a reconstructed data table, executable Python code, rendered output, visual style metadata, and an author-written finding selected from the experimental context. Three MLLMs independently propose chart classifications and metadata candidates, after which human experts separately curate rendered-chart fidelity and chart--finding consistency. This procedure yields 4,342 chart instances from 2,336 published papers.

SciChartAlign contributes:
\begin{itemize}[leftmargin=1.1em,itemsep=1pt,topsep=2pt,parsep=0pt]
\item \textbf{Published-document grounding}: 4,342 charts from 2,336 published papers rather than synthetic or web-scraped charts, with paper-native fields kept separate from reconstructed metadata so that users can audit which evidence originates in the publication.
\item \textbf{Scientific-evidence-aware alignment}: beyond caption- or QA-centered pairing, each chart is connected to the author-written finding that it supports in the experimental text.
\item \textbf{Reproducible evidence metadata}: each instance provides not only the chart image and text but also a reconstruction-oriented data table, executable plotting code, rendered output, and visual style attributes, so a chart can be re-rendered and inspected rather than only viewed.
\item \textbf{Bridge between understanding and generation}: the aligned fields serve as in-context examples for chart data extraction, chart explanation, and retrieval-augmented chart-to-code generation, for which we report reliability and utility evidence.
\end{itemize}

External rendered-chart selection reaches 0.812 accuracy and 0.701 Fleiss' $\kappa$; strict finding-selection accuracy is 0.641, but 83.19\% of non-selected candidate findings are semantically covered by the curated finding and 2.34\% partially covered. The aligned examples improve chart-to-table extraction across three models and caption-style explanation over data and rendering alone, and retrieving findings together with code attains the highest LLMScore (0.7624) and the lowest execution error rate (0.1295).

\begin{table*}[t]
\centering
\footnotesize
\setlength{\tabcolsep}{2.8pt}
\renewcommand{\arraystretch}{1}
\begin{tabular}{llccccccc}
\toprule
\multirow{2}{*}{\textbf{Dataset}}
& \multirow{2}{*}{\textbf{Source}}
& \multirow{2}{*}{\textbf{Chart}}
& \multirow{2}{*}{\textbf{Code}}
& \multirow{2}{*}{\textbf{Data Table}}
& \multicolumn{4}{c}{\textbf{Text Align.}} \\
\cmidrule(lr){6-9}
& & & & & \textbf{QA} & \textbf{Caption} & \textbf{Findings} & \textbf{Abstract} \\
\midrule
ChartQA~\cite{chartqa}             & Non-paper & \cmark & \xmark & \cmark & \cmark & \xmark & \xmark & \xmark \\
Chart-to-Text~\cite{Chart-to-Text} & Non-paper & \cmark & \xmark & \cmark & \xmark & \cmark & \xmark & \xmark \\
PlotQA~\cite{PlotQA}               & Non-paper & \cmark & \xmark & \cmark & \cmark & \xmark & \xmark & \xmark \\
ChartCoder~\cite{chartcoder}       & Non-paper & \cmark & \cmark & \pmark & \xmark & \xmark & \xmark & \xmark \\
CharXiv~\cite{charxiv}             & Preprint  & \cmark & \xmark & \xmark & \cmark & \xmark & \xmark & \xmark \\
SciGraphQA~\cite{scigraphqa}       & Preprint  & \cmark & \xmark & \pmark & \cmark & \xmark & \cmark & \cmark \\
SPIQA~\cite{spiqa}                 & Preprint  & \pmark & \xmark & \xmark & \cmark & \xmark & \xmark & \xmark \\
ChartMimic~\cite{chartmimic}       & Preprint  & \cmark & \cmark & \pmark & \xmark & \xmark & \xmark & \xmark \\
SCI-CQA~\cite{rethinking}          & Published & \cmark & \xmark & \xmark & \cmark & \cmark & \xmark & \xmark \\
ACL-Fig~\cite{ACL-Fig}             & Published & \pmark & \xmark & \xmark & \xmark & \cmark & \cmark & \xmark \\
\midrule
\textbf{SciChartAlign (ours)}      & Published & \cmark & \cmark & \cmark & \xmark & \cmark & \cmark & \xmark \\
\bottomrule
\end{tabular}
\caption{Comparison between SciChartAlign and existing chart-related datasets. Non-paper denotes a non-paper source, Preprint a publicly available preprint, and Published a conference- or journal-published paper. A component is marked \cmark{} when the resource releases it for every instance as a distinct field, \pmark{} when it is available for only part of the collection or must be derived rather than read from a released field, and \xmark{} when the resource does not provide it.}
\label{tab:dataset_comparison_arr}
\end{table*}

\section{Related Work}
\label{sec:related_work_arr}

\paragraph{Chart understanding and scientific figures.}
FigureQA, DVQA, and PlotQA provide synthetic chart-QA benchmarks, while ChartQA includes human-written questions over web-collected charts~\cite{figureqa,DVQA,PlotQA,chartqa}. CharXiv evaluates descriptive and reasoning questions over arXiv charts, and MultiChartQA targets reasoning across multiple charts~\cite{charxiv,multichartqa}. Scientific-document resources include ACL-Fig for figure-type classification and SPIQA for multimodal question answering over figures and tables~\cite{ACL-Fig,spiqa}.

\paragraph{Chart description and generation.}
Chart-to-Text and ChartSumm generate natural-language descriptions of trends, comparisons, and extrema in charts~\cite{Chart-to-Text,Rahman2023ChartSumm}. Visualization-generation research addresses the reverse direction: LIDA and Data Formulator generate visualizations from data or user intent, while MatPlotAgent and PlotGen use LLM-based agents for scientific visualization workflows~\cite{lida,Data-Formulator,matplotagent,PlotGen}.

\paragraph{Chart-to-code datasets.}
ChartMimic evaluates visually grounded code generation using scientific-paper charts, ChartCoder provides chart--code training pairs, and ChartX studies chart reasoning with aligned chart, data, and code modalities~\cite{chartmimic,chartcoder,chartX-and-chartVLM}.

Across these streams, supervision is organized around chart reasoning, textual description, or executable reconstruction, but not around the relation between a chart and the claim it supports. SciChartAlign combines published provenance, reconstructed data, executable code and output, visual style, captions, and author-written findings in one curated tuple; Table~\ref{tab:dataset_comparison_arr} presents this component-level positioning rather than a ranking of datasets.

\section{Chart-Message Aligned Dataset}
\label{sec:dataset_arr}

\begin{figure*}[t]
    \centering
    \includegraphics[width=0.96\linewidth]{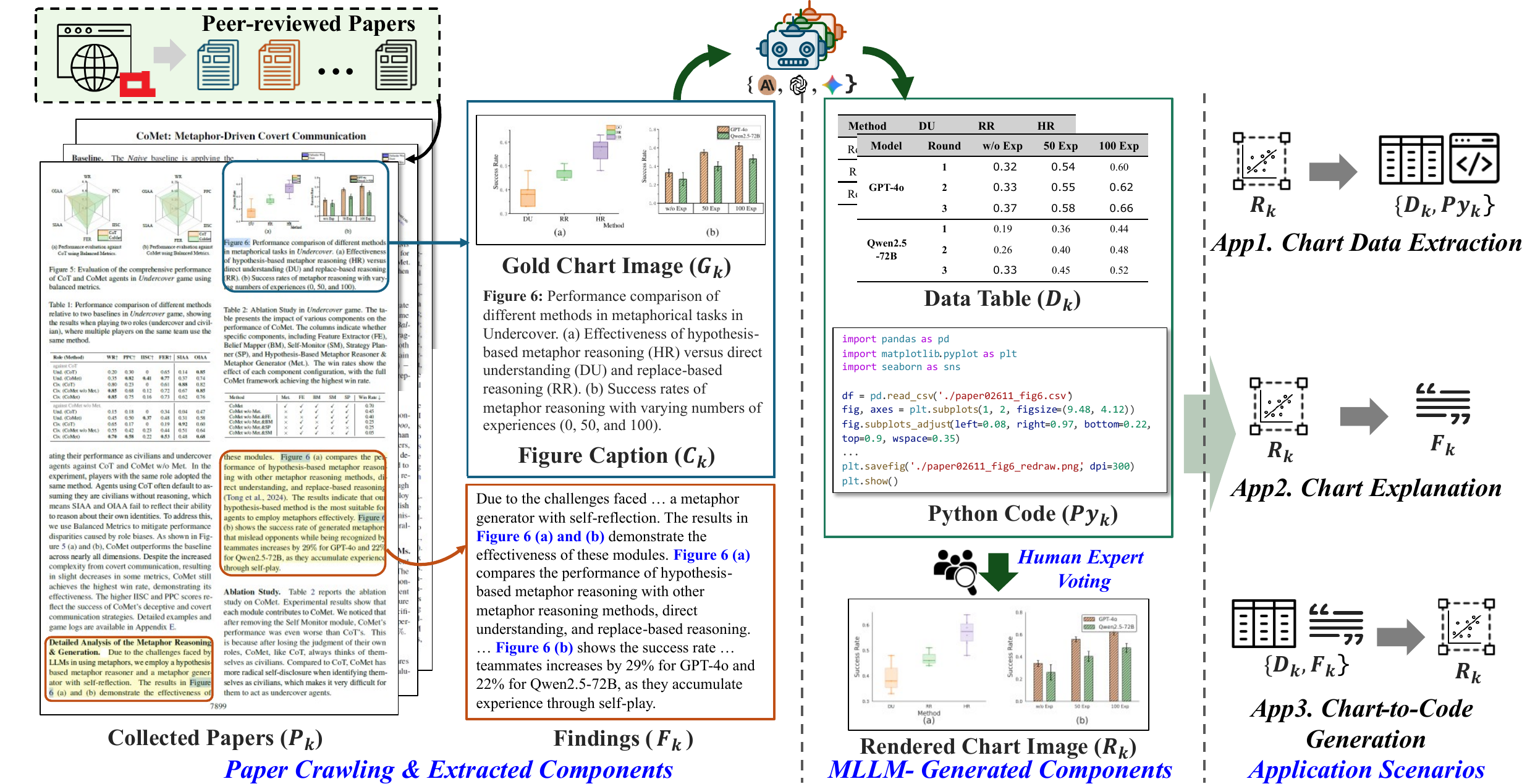}
    \caption{SciChartAlign construction. Published-paper charts and surrounding text are converted into aligned metadata candidates by three MLLMs; human curation selects visually faithful rendered charts and chart-consistent findings.}
    \label{fig:overall_arr}
\end{figure*}

\subsection{Schema and Design Goals}

For a validated chart instance $k$, the dataset tuple is
\begin{equation}
T_k=\{G_k,C_k,D_k,Py_k,S_k,F_k,R_k\},
\label{eq:tuple_arr}
\end{equation}
where $G_k$ and $C_k$ are the chart image and caption extracted from paper $P_k$; $D_k$ is a reconstruction-oriented data table; $Py_k$ is executable plotting code; $S_k$ records visual attributes; $F_k$ is an author-written finding; and $R_k$ is the execution result of $Py_k$. The figure-reference sentence and the extracted experimental section are retained alongside each tuple as document context: they ground the selection of $F_k$ and are released with the instance, but they are not themselves elements of $T_k$.

The tuple separates paper-native fields ($G_k$, $C_k$, $F_k$), which remain grounded in the published document, from reconstructed fields ($D_k$, $Py_k$, $S_k$, $R_k$), which make the chart computationally inspectable. The separation is recorded in the release so that users can distinguish author-provided content from metadata reconstructed during dataset creation.

The schema targets three requirements: \textit{document grounding} preserves the source paper, figure crop, caption, figure-reference sentence, and experimental context from which the chart--message relation is established; \textit{reconstructability} represents visible evidence through a structured table, executable program, rendered output, and explicit style attributes; and \textit{message alignment} links these artifacts to an author-written finding rather than a newly generated interpretation. A tuple is complete only when its visual reconstruction and finding alignment are both accepted during curation.

$D_k$ is an auditable reconstruction of the visible categories, series, values, and comparisons rather than the unavailable raw measurements, $R_k$ is not a pixel-identical copy of $G_k$, and $F_k$ preserves a supported statement without implying that a chart has only one valid interpretation.

\subsection{Source Collection and Native Component Extraction}

We collect 2025 main-track long papers from ACL, EMNLP, NAACL, and COLING. Experiment-, result-, analysis-, and discussion-related sections are parsed to locate figure references and nearby finding candidates. Three MLLMs distinguish data charts from other figures, and a candidate is retained when at least two models classify it as a chart. From 4,779 candidate papers, the pipeline retains 2,336 papers and 4,342 validated charts; a paper is dropped when no experiment-related section is detected or when no chart figure is referenced inside the detected sections. Detailed extraction rules and observed PDF failures are provided in Appendix~\ref{app:construction_arr}.

Figure-reference matching uses deterministic PDF preprocessing before MLLM-assisted filtering. Heading and keyword patterns identify experimental sections, their text is segmented into sentences and matched to explicit references such as ``Figure~3'' and ``Fig.~3,'' and layout information recovers the associated figure region and caption. Each retained crop is linked to its caption, its explicit reference sentence, and the surrounding experimental section, which is necessary because a reference sentence can merely direct the reader to a chart while the supported finding occurs in an adjacent sentence or paragraph. The procedure keeps the textual span used for finding identification separate from the cropped image used for reconstruction.

The three-model vote removes architecture diagrams, flowcharts, and qualitative examples without relying on a single classifier. Remaining failures are recorded; Appendix~\ref{app:construction_arr} lists the recurring section- and figure-level cases. Flagged crops are manually re-extracted when possible; otherwise the candidate is excluded before aligned metadata are generated.

\begin{table*}[t!]
\centering
\fontsize{10.5}{11}\selectfont
\setlength{\tabcolsep}{4.2pt}
\renewcommand{\arraystretch}{0.88}
\begin{tabular}{lrrrrrrrrr}
\toprule
\multirow{2}{*}{\textbf{Conf.}}
& \multirow{2}{*}{\textbf{Papers}}
& \multicolumn{8}{c}{\textbf{Chart Type}} \\
\cmidrule(lr){3-10}
& & \textbf{Bar} & \textbf{Line} & \textbf{Scatter} & \textbf{Heat/Mat.} & \textbf{Radar/Polar} & \textbf{Multi/Comp.} & \textbf{Others} & \textbf{Total} \\
\midrule
ACL    & 932 & 577 & 647 & 139 & 122 & 54 & 84 & 159 & \textbf{1,782} \\
EMNLP  & 903 & 539 & 583 & 136 & 134 & 54 & 86 & 131 & \textbf{1,663} \\
NAACL  & 275 & 178 & 161 & 50 & 41 & 20 & 24 & 41 & \textbf{515} \\
COLING & 226 & 126 & 134 & 36 & 26 & 11 & 12 & 37 & \textbf{382} \\
\midrule
\textbf{Total} & \textbf{2,336} & \textbf{1,420} & \textbf{1,525} & \textbf{361} & \textbf{323} & \textbf{139} & \textbf{206} & \textbf{368} & \textbf{4,342} \\
\bottomrule
\end{tabular}
\caption{Conference-level dataset statistics and chart-type distribution in SciChartAlign. Bar and Line include their grouped, stacked, horizontal, plot, and chart variants. Heat/Mat. denotes heatmap or matrix charts, Multi/Comp. denotes multi-panel or composite charts, and Others groups the remaining chart types.}
\label{tab:dataset_statistics_arr}
\end{table*}

\subsection{Aligned Metadata Construction}

As shown in Figure~\ref{fig:overall_arr}, Claude-, GPT-, and Gemini-family MLLMs receive the same input package---$G_k$, $C_k$, the reference sentence, and the extracted experimental context---and the same output requirements, with only API-specific formatting differences. Each model reconstructs $D_k$, generates $Py_k$ and $S_k$, and identifies candidate $F_k$ spans from the supplied paper text rather than generating new claims. Executing $Py_k$ yields $R_k$; if the program fails, the same model receives the execution error for one bounded correction while retaining the reconstructed data and intended visual structure. The prompt template is in Appendix~\ref{app:construction_arr}.

\paragraph{Reconstructed data and code.}
$D_k$ records axes, series, categories, legends, and visible values when they can be read from the chart. Bar and line charts preserve category order, series identity, legend mapping, and readable values; scatter plots preserve point coordinates when the axes permit recovery; heatmaps retain row--column structure and cell values. Distributional plots, regression fits, densities, and error bands frequently do not expose their latent observations, so for these cases $D_k$ stores summary values sufficient to reproduce the displayed aggregate pattern. $Py_k$ uses this explicit table with standard plotting libraries, and $R_k$ provides an executable check of the reconstruction rather than relying on code text alone.

\paragraph{Style and findings.}
$S_k$ records chart family, subplot organization, axis orientation and scale, color and marker encodings, line styles and hatching, grids, legend placement, and annotations. $F_k$ is an author-written span, ranging from a single sentence to a short passage, describing a trend, comparison, numerical relation, performance difference, or interpretation supported by the chart. A direct finding explicitly names the figure, whereas an indirect finding may occur in the same experimental discussion without a figure number; indirect candidates are retained only when their entities and relations can be traced to the chart and surrounding context.

For reproducibility, every candidate retains a full provenance record, itemized in Appendix~\ref{app:construction_arr}, so the release can be audited at the source, program-execution, rendering, and human-selection stages rather than only at the final tuple.

\subsection{Human Curation and Final Tuple Construction}

Three human experts review each candidate on two independent axes: rendered-chart selection requires preservation of the intended comparison rather than pixel identity, and finding selection asks whether the chosen text is supported by $G_k$ and its experimental context. Keeping the decisions separate avoids accepting a semantically plausible finding because its rendered candidate is visually strong, and only tuples passing both checks enter the final dataset.

The rendered-chart review applies three criteria. \textit{Style consistency} checks chart type, palette, markers, line styles, hatching, grids, and layout. \textit{Value equality} checks bar heights, point locations, line trends, and relative relationships that determine the visual comparison. \textit{Visual clarity} checks clipping, overlap, misalignment, and excessive density that could change interpretation.

For findings, when several candidates describe different valid aspects of the chart, experts select the span that best represents the principal message or merge complementary spans. This construction-stage curation is distinct from the external reliability study in Section~\ref{sec:experiments_arr}, which tests whether independent evaluators reproduce these choices. Full prompts and audit records are in Appendix~\ref{app:construction_arr}.

\subsection{Dataset Composition}

Table~\ref{tab:dataset_statistics_arr} reports conference-level composition and chart-family coverage: ACL, EMNLP, NAACL, and COLING contribute 1,782, 1,663, 515, and 382 charts from 2,336 papers. Bar and line charts account for 67.82\% of the collection, while five lower-frequency groups cover scatter, heat/matrix, radar/polar, and multi-panel or composite charts, together with a remaining group that includes distributional plots. Table~\ref{tab:metadata_statistics_arr} reports the scale of the aligned fields.

Bar and line charts offer explicit categories and ordered trends that are amenable to reconstruction, whereas scatter, radar/polar, heat/matrix, and composite charts more often depend on latent observations, angular order, dense cells, or multiple panels. The dataset retains these families, and the imbalance bounds the scope of the results reported below.

Findings are longer than captions because they often carry experimental conditions and interpretation, the median reconstructed table contains 51 rows and 240 cells, and the median program is 39 lines with 50 style fields. P95 values are reported because long experimental contexts and large tables create a tail that averages hide.

\begin{table*}[t]
\centering
\small
\setlength{\tabcolsep}{4.2pt}
\renewcommand{\arraystretch}{1.15}
\begin{tabular}{cc|ccc|ccc}
\toprule
\textbf{Evaluator} & \textbf{Target}
& \textbf{Accuracy $\uparrow$}
& \textbf{Cohen's $\kappa$ $\uparrow$}
& \textbf{Fleiss' $\kappa$ $\uparrow$}
& \textbf{JSD $\downarrow$}
& \textbf{$p$-value}
& \textbf{Cramer's $V$ $\downarrow$} \\
\midrule
\multirow{2}{*}{\makecell[c]{Human\\Expert}}
& Render Chart Voting & 0.812 & 0.506 & 0.701 & 0.021 & $<.001$ & 0.118 \\
\cmidrule{2-8}
& Finding Voting      & 0.641 & 0.326 & 0.382 & 0.024 & $<.001$ & 0.126 \\
\cmidrule{1-8}
\multirow{2}{*}{\makecell[c]{LLM-as-a\\Judge}}
& Render Chart Voting & 0.786 & 0.458 & 0.668 & 0.025 & $<.001$ & 0.105 \\
\cmidrule{2-8}
& Finding Voting      & 0.603 & 0.281 & 0.334 & 0.028 & $<.001$ & 0.112 \\
\bottomrule
\end{tabular}
\caption{Voting reliability for human and LLM-as-a-Judge evaluators. Accuracy and Cohen's $\kappa$ are computed against the curated label; Fleiss' $\kappa$ is computed among the external evaluators. JSD and Cramer's $V$ measure the shift between the construction and external selection distributions, and $p$ is from a $\chi^2$ test of independence on those distributions.}
\label{tab:voting_reliability_arr}
\end{table*}

\begin{figure*}[t]
  \centering
  \begin{subfigure}[t]{0.49\textwidth}
    \centering
    \includegraphics[width=\linewidth]{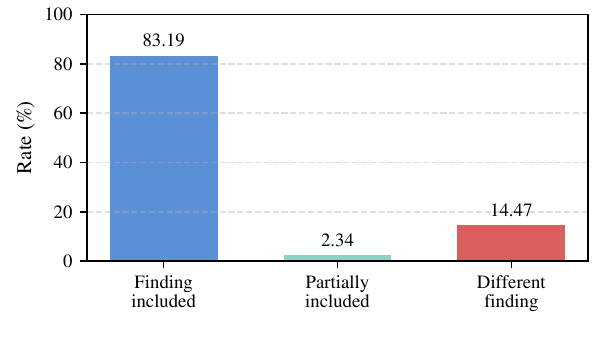}
    \caption{Semantic coverage}
  \end{subfigure}\hfill
  \begin{subfigure}[t]{0.49\textwidth}
    \centering
    \includegraphics[width=\linewidth]{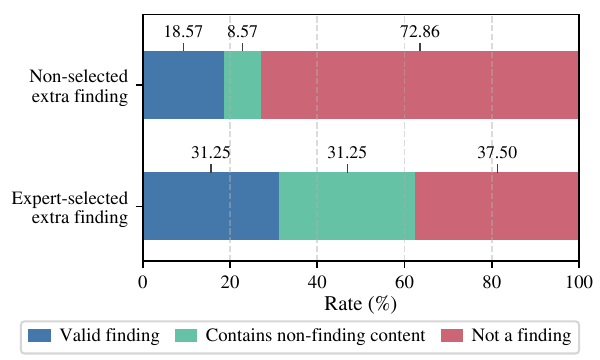}
    \caption{Validity of additional findings}
  \end{subfigure}
  \caption{Finding audit on $\mathcal{S}_{400}$. (a) coverage of the non-selected candidate findings by the curated finding; (b) validity of the content those candidates add.}
  \label{fig:finding_validity_arr}
\end{figure*}

\section{Applications}
\label{sec:applications_arr}

The aligned fields support three applications.

\subsection{Chart Data Extraction}

Chart data extraction converts marks, axes, legends, and labels into a structured table. SciChartAlign provides $(G_k,C_k,D_k)$ examples from published papers, including series and categories that do not follow a fixed synthetic schema. $D_k$ remains reconstruction-oriented: success means recovering the visible structure and value pattern, not recovering latent raw measurements that are absent from the raster image.

This application tests whether dataset examples are useful as in-context demonstrations: a model receives a target chart and caption under zero-shot, random few-shot, or same-type few-shot conditions, so improvement over zero-shot measures the utility of the aligned tables and the gap between retrieval conditions reveals whether chart family alone suffices for selecting demonstrations.

\subsection{Chart Explanation}

Chart explanation must cover both what is plotted and what message the chart supports. We compare a reconstructed table and chart alone ($D+R$), the same inputs with chart-observable generic guidance ($D+R+GI$), and inputs augmented with the author-written finding ($D+R+F$). The task generates a caption-style explanation against the published caption.

The comparison contrasts two ways of adding textual context to the same visual input. $GI$ is generated for each chart from the chart image together with $C_k$, $S_k$, and $D_k$, under an explicit constraint to state only observable structure---chart family, axis semantics, scales and units, and the purpose of visible annotations---and never to assert a paper-level claim. It is a structural instruction derived from the aligned metadata rather than a restatement of the caption. $F$ supplies the author-written statement selected from the paper. Comparing them tests whether either context contributes information beyond data and rendering.

\subsection{Chart-to-Code Generation}

Chart-to-code generation uses aligned examples to produce executable plotting programs. Each target supplies a data table and generation instruction; retrieval returns examples containing a chart, text metadata, and code.

We compare QA-, caption-, code-, mention-, and finding-plus-code retrieval sources. Every condition uses the same similarity-based retriever, the same number of retrieved examples, and the same generation instruction, so the only manipulated factor is the corpus from which the few-shot examples are drawn. This application measures both program validity and message-aware output quality rather than code-text plausibility alone.

\begin{table}[t!]
\centering
\footnotesize
\setlength{\tabcolsep}{2.2pt}
\renewcommand{\arraystretch}{1.1}
\begin{tabular*}{\columnwidth}{@{\extracolsep{\fill}}llrrrr}
\toprule
\textbf{Model} & \textbf{Cond.} & \textbf{Parse} & \textbf{F1} & \textbf{Value} & \textbf{Comb.} \\
\midrule
~ & Zero-shot & 1.000 & 0.1009 & 0.1250 & 0.1105 \\
Gemini & Random FS & 0.875 & 0.1749 & 0.3704 & 0.2531 \\
~ & Same-type FS & 1.000 & 0.1290 & 0.2494 & 0.1772 \\
\midrule
~ & Zero-shot & 1.000 & 0.1414 & 0.1250 & 0.1349 \\
GPT & Random FS & 1.000 & 0.1923 & 0.5343 & 0.3291 \\
~ & Same-type FS & 1.000 & \textbf{0.2099} & \textbf{0.6588} & \textbf{0.3895} \\
\midrule
~ & Zero-shot & 1.000 & 0.0351 & 0.0914 & 0.0576 \\
Claude & Random FS & 1.000 & 0.1010 & 0.3745 & 0.2104 \\
~ & Same-type FS & 1.000 & 0.0658 & 0.3742 & 0.1892 \\
\bottomrule
\end{tabular*}
\caption{Data-table extraction from chart images. FS denotes few-shot; Parse denotes the share of outputs parsed as a table; F1 denotes table-structure F1; Value denotes value similarity; Comb. denotes $0.6$ F1 $+0.4$ Value.}
\label{tab:data_extraction_results_arr}
\end{table}

\section{Experiments}
\label{sec:experiments_arr}

\subsection{Evaluation Setup}

We first validate whether external evaluators reproduce the curated rendered-chart and finding choices, then evaluate the three applications. Repeated human comparison and generation experiments use a 400-chart subset $\mathcal{S}_{400}$ drawn from the 4,342 validated instances. At $n=400$ and $N=4{,}342$, a proportion measured on this subset carries a conservative 95\% finite-population margin of error of approximately 4.7 percentage points, which we attach to the reliability and preference rates below. The chart-to-table pilot uses eight targets from this subset and evaluates Gemini, GPT, and Claude. Subset composition, metric definitions, prompts, model settings, and retrieval details are in Appendix~\ref{app:experiments_arr}.

Rendered-chart and finding voting report accuracy and Cohen's $\kappa$ against the curated label, and Fleiss' $\kappa$ among the external evaluators. Jensen--Shannon divergence (JSD) and Cramer's $V$ measure evaluator preference shift, and a $\chi^2$ test of independence assesses whether the two evaluator groups differ in selection distribution. Application metrics follow the output type: structure F1 and value similarity for tables; blind three-way preference, BERTScore, and ROUGE-L for explanations; and LLMScore, execution error rate, and pairwise preference for chart-to-code generation.

\paragraph{Reliability protocol.}
External evaluators repeat two three-way choices without access to the original curation decision: rendered-chart voting compares $G_k$ with three candidate $R_k$ images under the value, style, layout, and clarity criteria used in construction, and finding voting selects the candidate that best represents the chart-supported message. Because several findings can be valid for one chart, strict agreement is supplemented by a semantic coverage audit.

\paragraph{Application protocols.}
The extraction model emits a free-form table, which is parsed after decoding to avoid hiding grouping and schema failures behind a fixed answer format. Chart-to-code retrieval returns three examples, and every generated program is executed; failure to produce a saved figure contributes to the reported error rate.

\subsection{Dataset Reliability}

Table~\ref{tab:voting_reliability_arr} shows higher reproducibility for rendered-chart selection than for finding selection. Human evaluation obtains 0.812 accuracy and 0.701 Fleiss' $\kappa$ for rendered charts, compared with 0.641 and 0.382 for findings; LLM judging follows the same ordering. For rendered charts, agreement among the external evaluators exceeds their agreement with the curated label (Fleiss' $\kappa$ 0.701 versus Cohen's $\kappa$ 0.506), indicating that the evaluators apply a stable shared criterion while the curated label is one admissible choice under that criterion rather than the only one. JSD (0.021--0.028) and Cramer's $V$ (0.105--0.126) remain low even though the $\chi^2$ tests reach $p<.001$: at $n=400$ the tests detect a reliable but small shift, so the two groups differ in how strongly they prefer particular candidates rather than in which candidates they accept.

The gap between the two tasks is consistent across human and LLM evaluation: finding accuracy is lower by 0.171 for humans and 0.183 for the LLM judge. A finding label is therefore not a uniquely determined caption substitute: the dataset provides one curated, chart-grounded message, and the coverage audit records how often independent choices fall within its semantic scope.

The audit covers all 400 charts of $\mathcal{S}_{400}$. For each chart, three MLLMs propose a finding candidate and expert voting selects one of them as the curated finding; the two non-selected candidates are then examined for whether their content is already contained in that curated finding. Under this protocol the curated finding covers 83.19\% of the non-selected candidates and partially covers 2.34\%. Among the content those candidates add beyond the curated finding, 31.25\% is itself a valid chart-supported finding and another 31.25\% mixes finding and non-finding material. Figure~\ref{fig:finding_validity_arr} separates semantic coverage from strict label agreement.

\begin{table}[t!]
\centering
\footnotesize
\setlength{\tabcolsep}{3.4pt}
\renewcommand{\arraystretch}{1.08}
\begin{tabular*}{\columnwidth}{@{\extracolsep{\fill}}lrrrr}
\toprule
\textbf{Input} & \textbf{Human} & \textbf{LLM} & \textbf{BERT} & \textbf{R-L} \\
\midrule
Baseline ($D{+}R$) & 0.1575 & 0.1525 & 0.7137 & 0.1262 \\
~~$+\,GI$ & \textbf{0.4975} & \textbf{0.4725} & \textbf{0.7481} & \textbf{0.1788} \\
~~$+\,F$  & 0.3450 & 0.3750 & 0.7393 & 0.1645 \\
\bottomrule
\end{tabular*}
\caption{Caption-style chart explanation. Baseline is the default input, the reconstructed data table $D$ with the rendered chart $R$; the other rows add the generic instruction $GI$ or the author-written finding $F$ to that input. Human and LLM denote three-way selection rates; BERT and R-L denote BERTScore and ROUGE-L.}
\label{tab:caption_generation_results_arr}
\end{table}

\begin{table}[t]
\centering
\footnotesize
\setlength{\tabcolsep}{2.6pt}
\renewcommand{\arraystretch}{1.08}
\begin{tabular*}{\columnwidth}{@{\extracolsep{\fill}}llrrr}
\toprule
\textbf{Source} & \textbf{Mod.} & \textbf{Win} & \textbf{LLM} & \textbf{Err.} \\
\midrule
ChartQA & QA & 0.5763 & 0.6572 & 0.2384 \\
Chart2Text & Caption & 0.5488 & 0.6819 & 0.2147 \\
Chart2Code & Code & 0.5131 & 0.7048 & 0.1763 \\
SciGraphQA & Mention & 0.5615 & 0.6645 & 0.2268 \\
\textbf{Ours} & $F+$Code & -- & \textbf{0.7624} & \textbf{0.1295} \\
\bottomrule
\end{tabular*}
\caption{Few-shot source comparison for retrieval-augmented chart-to-code generation. Mod. denotes input modality; Win denotes the blind pairwise win rate of $F+$Code over the source in that row, with no self-comparison for the reference row; LLM denotes LLMScore; Err. denotes code-generation error rate.}
\label{tab:retrieval_code_generation_arr}
\end{table}

\subsection{Application Results}

\paragraph{Chart data extraction.}
Table~\ref{tab:data_extraction_results_arr} shows that both few-shot conditions improve the combined score over zero-shot for all three models, with GPT and same-type examples highest (0.3895). Random examples outperform same-type examples for Gemini and Claude, so chart type alone is not a uniformly effective retrieval criterion, and because the pilot contains eight targets with modest absolute scores it demonstrates few-shot utility rather than recovery of original measurements. Structure remains the bottleneck: models often recover plausible values while renaming columns, reshaping long-form tables to wide form, or merging categories, and the metric penalizes these transformations because $D_k$ must preserve the visible series, axis, legend, and category relations.

\paragraph{Chart explanation.}
Table~\ref{tab:caption_generation_results_arr} shows that both added-context conditions improve over the baseline. Generic instruction obtains the highest Human preference (0.4975), LLM preference (0.4725), BERTScore (0.7481), and ROUGE-L (0.1788); the author-written finding ranks second on all four. The ordering follows the target of the task: a published caption states chart family, axes, units, and plotting conditions, which is precisely the structural content $GI$ is constrained to encode, whereas $F$ contributes an experimental claim that a caption frequently does not carry, and $GI$ is itself derived in part from $C_k$. The result therefore does not indicate that author-written findings are uninformative; it indicates that structural guidance is the closer match when the reference output is a caption. The chart-to-code results show the complementary case, in which the finding condition preserves the intended message.

\paragraph{Chart-to-code generation.}
Table~\ref{tab:retrieval_code_generation_arr} compares retrieved few-shot sources under a fixed retriever, a fixed number of retrieved examples, and a fixed generation instruction, so the manipulated factor is the source corpus rather than the retrieval algorithm. The finding-plus-code condition obtains the highest LLMScore (0.7624) and lowest execution error rate (0.1295); code-only is the strongest baseline (0.7048 and 0.1763). $F+$Code is also preferred by a majority in blind pairwise comparison against every source, from 0.5131 over code-only to 0.5763 over QA examples, and the ordering of these relative preferences matches the LLMScore and error-rate ordering.

Each source is compared as it is released, so a row bundles the modality it supplies with the provenance of its charts; the result identifies which existing resource forms the more useful demonstration pool, not, within this table, the contribution of findings apart from their published-paper origin. Appendix~\ref{app:qualitative_arr} adds a within-corpus qualitative comparison that varies only the added context: generating the same target from $D$, $D+C$, $D+C+GI$, and $D+C+F$ shows that caption and structural guidance recover style consistency, whereas only the finding condition also preserves the semantic relation the chart conveys. Retrieval details are in Appendix~\ref{app:experiments_arr}; generalization to other retrievers, plotting libraries, and chart domains remains untested.

\section{Conclusion}

We introduced SciChartAlign, a dataset that aligns published scientific charts with reconstructed data, executable code, rendered output, visual style, and author-written findings. External evaluation finds rendered-chart choices reproducible and finding labels semantically well covered despite lower strict agreement. Across three applications, the aligned examples improve chart-to-table extraction, added metadata context improves caption-style explanation over data and rendering alone, and finding-plus-code retrieval yields the highest LLMScore and lowest error rate in the evaluated chart-to-code setting. Broader domains and underrepresented chart families remain to be tested.

\clearpage
\section{Limitations}

SciChartAlign is limited to 2025 published NLP papers from ACL, EMNLP, NAACL, and COLING. Bar and line charts dominate the dataset, so results are less conclusive for non-Cartesian, distributional, and composite charts. Reconstructed tables preserve visible evidence but cannot recover latent raw observations from raster images. Finding selection is also semantically ambiguous: several statements can be valid for the same chart even when only one is stored as the curated label. The chart-to-table evaluation additionally inherits a label dependency: $D_k$ is produced by the same three MLLM families that are evaluated in Table~\ref{tab:data_extraction_results_arr}, so formatting conventions shared between a model and the reference table may raise its score independently of extraction ability, and the reported scores are best read as relative comparisons across prompting conditions rather than as absolute extraction accuracy. Finally, the application results characterize the evaluated models, prompts, subset, and retrieval procedure rather than all scientific-chart settings.

\IfFileExists{custom.bib}{
  \bibliography{custom}
}{
  \bibliography{latex/custom}
}

\appendix

\section{Dataset Construction Details}
\label{app:construction_arr}

\subsection{Extraction and Failure Handling}

Experiment-related sections are identified with heading patterns including \textit{Experiment}, \textit{Results}, \textit{Analysis}, and \textit{Discussion}. Section detection fails for approximately 8.2\% of candidate papers because of non-standard headings, complex layouts, merged headings, or PDF parsing errors. Other recurring failures include multi-subfigure references, implicit references without figure numbers, detached captions, and crops that omit panels or include surrounding text. Flagged figures are re-extracted manually when possible.

\begin{table}[t!]
\centering
\footnotesize
\renewcommand{\arraystretch}{1.08}
\setlength{\tabcolsep}{2.4pt}
\begin{tabular*}{\columnwidth}{@{\extracolsep{\fill}}p{0.35\columnwidth}rrrl}
\toprule
\textbf{Component} & \textbf{Unit} & \textbf{Avg.} & \textbf{Med.} & \textbf{P95}  \\
\midrule
Caption $C_k$        & words & 31.0 & 26.0 & 70.0\\
Finding $F_k$        & words & 139.8 & 114.0 & 331.9\\
Figure-ref. sentence  & words & 31.8 & 24.0 & 87.0\\
Experimental context  & words & 1730.1 & 1643.0 & 2810.0\\
Style $S_k$          & fields & 61.5 & 50.0 & 130.0\\
\midrule
Data table $D_k$     & cells & 432.6 & 240.0 & 1359.5\\
~    & rows & 83.8 & 51.0 & 251.9 \\
Python $Py_k$       & lines & 40.6 & 39.0 & 62.0 \\
~       & tokens & 475.5 & 441.0 & 808.9 \\
\bottomrule
\end{tabular*}
\caption{Metadata statistics of SciChartAlign, computed over all 4,342 validated chart instances.
Component lists the fields of the aligned tuple together with the document context retained alongside it; \textbf{Avg.}, \textbf{Med.}, and \textbf{P95} report the mean, the median, and the 95th percentile of each component's size across instances, measured in the unit given in the \textbf{Unit} column.}
\label{tab:metadata_statistics_arr}
\end{table}

\subsection{MLLM Prompt and Audit Records}

All three MLLMs receive the same logical instruction: given $G_k$, $C_k$, the figure-reference sentence, and experimental context, return $D_k$, executable $Py_k$, $S_k$, and a candidate author-written $F_k$. Outputs must not introduce values or claims unsupported by the chart or text. If execution fails, the same model receives the error log for one bounded correction. Candidate logs retain the paper identifier, source span, crop, caption, MLLM source, decoding configuration, execution status, error log, rendered path, and curation decision.

\begin{table*}[t]
\centering
\footnotesize
\renewcommand{\arraystretch}{1.10}
\begin{tabular}{p{0.115\linewidth}p{0.82\linewidth}}
\toprule
\textbf{Field} & \textbf{Content} \\
\midrule
Role instruction & Act as an expert in scientific visualization and experimental-result interpretation, and construct metadata required for chart reconstruction and paper-grounded interpretation. \\
Input package & Gold chart image $G_k$, caption $C_k$, figure-reference sentence, and extracted experimental-section text. \\
Required outputs & Reconstructed data table $D_k$, executable Python code $Py_k$, visual style metadata $S_k$, and candidate author-written findings $F_k$. \\
Output constraints & Use explicit data tables and executable rendering code, and avoid values or claims that are not grounded in the chart or paper text. \\
Correction policy & If code execution fails, revise the code using the execution error log within a bounded correction step while preserving the reconstructed data and intended visual structure. \\
\bottomrule
\end{tabular}
\caption{Logical prompt template used for MLLM-assisted metadata construction. The same template is used across MLLMs, with only API-specific formatting changes.}
\label{tab:prompt_template_arr}
\end{table*}

\subsection{Human Selection Criteria}

Rendered candidates are compared on chart type and layout, value relations, style attributes, and visual clarity. Finding candidates must describe a chart-observable trend, comparison, numerical relation, performance difference, or interpretation grounded in the paper context. Experts may merge complementary candidates when they jointly express the principal chart-supported message.

\section{Experimental Details}
\label{app:experiments_arr}

\subsection{Experimental Subset}

The 400-chart subset has the chart-type composition reported in Table~\ref{tab:experimental_sample_distribution_arr}. Under the worst-case variance assumption $p(1-p)=0.25$, its finite-population margin of error is
\begin{equation}
1.96\sqrt{\frac{0.25}{400}}\sqrt{\frac{4{,}342-400}{4{,}342-1}}\approx0.047.
\end{equation}
This bound governs the proportions measured on the subset, such as the voting accuracies in Table~\ref{tab:voting_reliability_arr} and the preference rates in Tables~\ref{tab:caption_generation_results_arr} and~\ref{tab:retrieval_code_generation_arr}. Table~\ref{tab:experimental_sample_distribution_arr} separately reports the chart-type composition of $\mathcal{S}_{400}$ itself: relative to the full distribution in Table~\ref{tab:dataset_statistics_arr}, the subset contains a higher share of bar charts and a lower share of scatter, radar/polar, and composite charts. That composition describes the evaluated sample and is not put forward as an estimate of the population composition.

\begin{table}[t]
\centering
\footnotesize
\setlength{\tabcolsep}{4pt}
\renewcommand{\arraystretch}{1.08}
\begin{tabular*}{\columnwidth}{@{\extracolsep{\fill}}lrr}
\toprule
\textbf{Chart Type} & \textbf{Count} & \textbf{Share (\%)} \\
\midrule
Bar & 198 & 49.50 \\
Line & 133 & 33.25 \\
Scatter & 15 & 3.75 \\
Heatmap & 23 & 5.75 \\
Radar/Polar & 3 & 0.75 \\
Multi/Composite & 9 & 2.25 \\
Others & 19 & 4.75 \\
\midrule
\textbf{Total} & \textbf{400} & \textbf{100.00} \\
\bottomrule
\end{tabular*}
\caption{Chart-type distribution of the experimental subset $\mathcal{S}_{400}$. Count is the number of instances of each family in $\mathcal{S}_{400}$; Share is that family's percentage of the 400 sampled charts.}
\label{tab:experimental_sample_distribution_arr}
\end{table}

\subsection{Task Protocols and Metrics}

For reliability, three external human experts and an LLM judge repeat rendered-chart and finding voting. For extraction, zero-shot, random few-shot, and same-type few-shot conditions generate free-form tables; the combined score is $0.6F1_{\mathrm{str}}+0.4\mathrm{Val}$. For explanation, $D+R$, $D+R+GI$, and $D+R+F$ generate captions evaluated by blind three-way preference, BERTScore~\cite{zhang2020bertscore}, and ROUGE-L~\cite{lin2004rouge}. For chart-to-code, a TF--IDF seed index and chart--entity graph retrieve three examples; the identical retriever and index-construction procedure are applied to every candidate source corpus, so conditions differ only in the corpus being indexed. Generated programs must execute and call \texttt{savefig}; outputs are evaluated by LLMScore, error rate, and blind pairwise preference of $F+$Code over each retrieved source. Model versions, prompts, decoding parameters, library versions, execution scripts, and environment files are provided in the public repository.

\twocolumn[{%
\section{Qualitative Chart-Generation Example}
\label{app:qualitative_arr}
\vspace{0.8em}
\begin{center}
  \includegraphics[width=\textwidth]{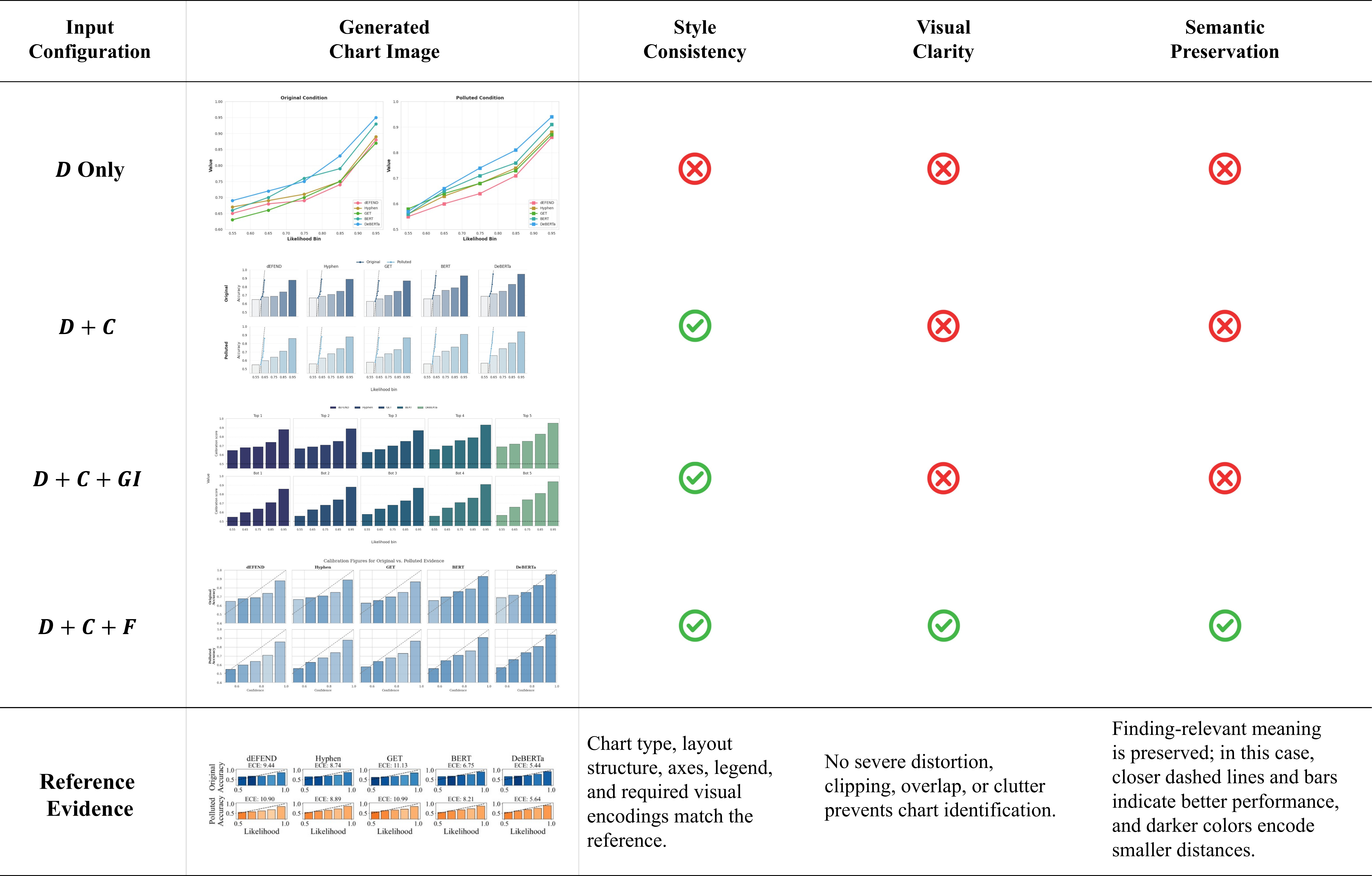}
  \captionof{figure}{Qualitative chart-generation comparison for the chart-to-code setting. The same target is generated from four input configurations and each result is checked for style consistency, visual clarity, and semantic preservation against the reference evidence. $D$, $C$, $GI$, and $F$ denote data table, caption, generic instruction, and finding.}
  \label{fig:quality_chart_generation_arr}
\end{center}
}]

\end{document}